%% file: egpaper.tex
\documentclass[10pt,twocolumn,letterpaper]{article}

\usepackage{wacv}
\usepackage{times}
\usepackage{epsfig}
\usepackage{graphicx}
\usepackage{amsmath}
\usepackage{amssymb}
\usepackage[T1]{fontenc}
\usepackage{breqn}
\usepackage{comment}

\usepackage{pifont}
\newcommand{\cmark}{\ding{51}}%
\usepackage{tablefootnote}
\usepackage{threeparttable}
\usepackage{multirow}
\usepackage{arydshln}
\usepackage{stfloats}

\def\wacvPaperID{141} 

\wacvfinalcopy 

\ifwacvfinal
\def\assignedStartPage{1} 
\fi

\ifwacvfinal
\usepackage[breaklinks=true,bookmarks=false]{hyperref}
\else
\usepackage[pagebackref=true,breaklinks=true,colorlinks,bookmarks=false]{hyperref}
\fi

\ifwacvfinal
\else
\fi

\begin{document}

\title{Synthetic Expressions are Better Than Real for Learning to Detect Facial Actions}

\author{Koichiro Niinuma$^1$, Itir Onal Ertugrul$^2$, Jeffrey F Cohn$^3$, L{\'a}szl{\'o} A Jeni$^4$\\
$^1$Fujitsu Laboratories of America, Pittsburgh, PA, USA, $^2$Tilburg University, Tilburg, The Netherlands\\
$^3$University of Pittsburgh, Pittsburgh, PA, USA, $^4$Carnegie Mellon University, Pittsburgh, PA, USA\\

{\tt\small kniinuma@fujitsu.com, i.onal@uvt.nl, jeffcohn@pitt.edu, laszlojeni@cmu.edu}
}

\maketitle

\begin{abstract}
Critical obstacles in training classifiers to detect facial actions are the limited sizes of annotated video databases and the relatively low frequencies of occurrence of many actions. To address these problems, we propose an approach that makes use of facial expression generation. Our approach reconstructs the 3D shape of the face from each video frame, aligns the 3D mesh to a canonical view, and then trains a GAN-based network to synthesize novel images with facial action units of interest. To evaluate this approach, a deep neural network was trained on two separate datasets:  One network was trained on video of synthesized facial expressions generated from FERA17; the other network was trained on unaltered video from the same database. Both networks used the same train and validation partitions and were tested on the test partition of actual video from FERA17. The network trained on synthesized facial expressions outperformed the one trained on actual facial expressions and surpassed current state-of-the-art approaches. 
\end{abstract}

\section{Introduction}

Facial expressions convey emotional states, behavioral intentions, and physical state \cite{tian2001recognizing}. In behavior sciences the gold-standard to decode such facial expressions is the Facial Action Coding System (FACS) \cite{ekman2002facial}. FACS decomposes facial expressions into anatomically based action units (AUs), which alone or in combinations can represent nearly all possible facial expressions. While much progress has been made in action unit detection, at least two significant problems impede further advances.

\begin{table*}[]
\centering
\caption{Comparison of AU intensity datasets}
\label{table:datasets}
\begin{threeparttable}
\begin{tabular}{|l|ccccc|}
\hline
 & \begin{tabular}[c]{@{}c@{}}\# of AUs \\ with intensity codes\end{tabular}  & \begin{tabular}[c]{@{}c@{}}Continuous\\Video\end{tabular} & \begin{tabular}[c]{@{}c@{}}Manual \\ Ground Truth\end{tabular} & \begin{tabular}[c]{@{}c@{}}Social \\ Context\end{tabular} & 
 \begin{tabular}[c]{@{}c@{}}Manual Coding \\ Reliability\end{tabular}\\ \hline
 
FERA 2017~\cite{Valstar17}\tnote{i} & 7 & \cmark & \cmark & \cmark  & Good       \\
DISFA~\cite{Mavadati13} & 12 &  \cmark & \cmark & & Good          \\
EmotioNet~\cite{Quiroz16} & 12\tnote{ii} & & Semi automated\tnote{iii} & & Unknown            \\
UNBC Pain~\cite{Lucey11} & 10 & \cmark & \cmark &  & Good \\ 
\hline
\end{tabular}
\begin{tablenotes}\footnotesize
\item[i] The FERA 2017 dataset consists of BP4D~\cite{Zhang14} and BP4D+~\cite{Zhang16}.
\item[ii] In the EmotioNet Challenge 2020, 11 more AUs have been added for a total of 23 AUs. http://cbcsl.ece.ohio-state.edu/enc-2020/
\item[iii] They manually FACS-coded 10\% of this database.
\end{tablenotes}
\end{threeparttable}
\end{table*}
\label{tb:comp_baselines}

First, while much video of spontaneous facial expression has been collected, only a fraction have been manually annotated.  Expert FACS annotators are relatively few and the time required to comprehensively annotate action units slows the effort. AU annotation of a single minute of video typically requires one to three hours from a highly trained expert \cite{cohn2005measuring}. Reaching human-like accuracy in a fully supervised way would require labeled datasets orders of magnitude larger than those available today. 

Second, AU labels are highly skewed in spontaneous behavior. Many AUs occur rarely and only a sparse subset of AU intensities occur at a time. As a consequence, rare classes do not contribute equally during classifier training, which hinders learning and undermines global performance. Although imbalanced learning has been well explored in the past, most approaches deal with a single majority and minority class and are not directly applicable to the multi-label AU domain.

We propose a generative semi-supervised method that can handle within a common framework both the limited sizes of annotated data now available and the low frequencies of occurrence.

Our approach makes use of a 3D facial expression generator trained on the labeled portion. The approach first reconstructs the 3D shape of the face from each video frame, aligns the reconstructed meshes to a canonical view to establish semantic correspondence across frames and subjects, and then trains a GAN-based network to synthesize novel images with facial action units of interest. We then use the network to generate a large AU-balanced dataset from unlabeled images for training.

Intensity of facial actions may be one of the most important features in assessing a person’s emotional state ~\cite{mckeown2015gender}. Low intensity actions are detectable through motion~\cite{Ambadar2005}. Table~\ref{table:datasets} compares AU intensity datasets. In order to be able to detect these fine scale changes, we selected FERA 2017~\cite{Valstar17}, DISFA~\cite{Mavadati13} and UNBC Pain~\cite{Lucey11}, which are video datasets having manual fine-grained AU intensity annotations, to evaluate our approach (see supplementary materials for results on UNBC Pain). Note that Table~\ref{table:datasets} does not include datasets without AU intensity annotation, such as Aff-Wild2~\cite{kollias2019expression}.

Our novelties are twofold:

\textbf{3D geometry based AU manipulation.}
Unlike previous work on facial AU manipulation that is limited to either 2D representations \cite{Pumarola_ijcv2019} or individual frames \cite{geng20193d}, our approach uses the 3D structure of the face to create semantic correspondence across video-frames and subjects. 

\textbf{Synthetic multi-label stratification of AUs.}
Many AU occur infrequently, which undermines learning.  To avoid imbalanced classes, we increase the prevalence and variety of under-represented AUs by synthesizing new facial expression.

\section{Related Work}

\textbf{Solutions to limited AU annotations:} 
While massive amount of facial expression data is available, high quality annotations of AU intensity labels are limited. To mitigate the problems in the AU annotations, weakly-supervised, semi-supervised, and self-supervised approaches have been proposed. Weakly-supervised approaches aim to exploit incomplete, inaccurate or inexact annotations to provide supervision.  Zhao et al. \cite{Zhao18} proposed a weakly supervised clustering approach utilizing a large set of web images with inaccurate annotations. The annotations were obtained from either pretrained models or query strings. Ruiz et al. \cite{ruiz2015emotions} proposed to train AU detectors without any AU annotations by leveraging the expression labels and using prior knowledge on expression-dependent AU probabilities. Similarly, Zhang et al. \cite{zhang2018classifier} exploited expression-dependent and expression-independent joint AU probabilities as prior knowledge and learned to detect AUs without any AU annotation. In another study, Zhang et al. \cite{Zhang18} used various types of domain knowledge including relative appearance similarity, temporal intensity ordering, facial symmetry, and contrastive appearance difference to provide weak supervision for AU intensity estimation with extremely limited annotations. Peng et al. \cite{Peng18} proposed a method that learns AU classifiers from domain knowledge and expression-annotated facial images through adversarial training.

\begin{figure*}[!t]
\begin{center}
\includegraphics[width=.95\linewidth]{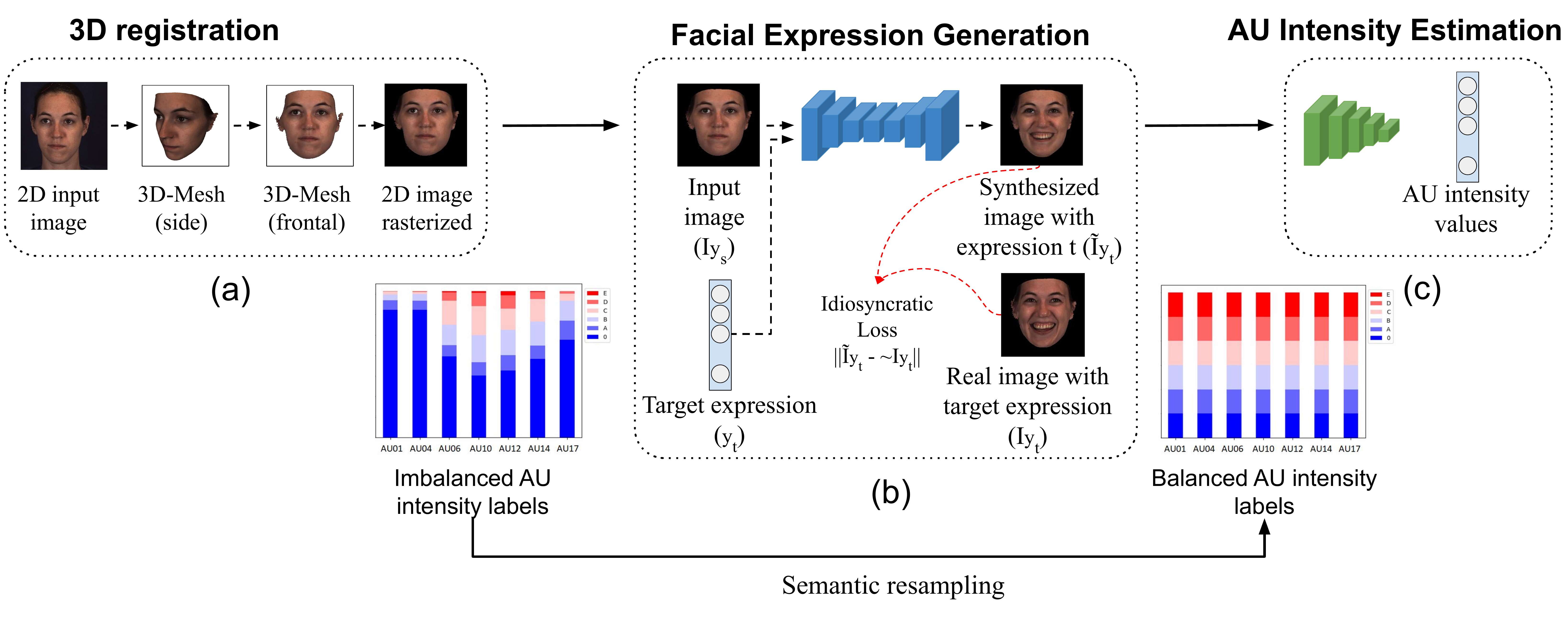}
\end{center}
\caption{Pipeline of our approach. (a) A dense 3D mesh of the face is reconstructed, rotated to the canonical frontal view and 2D rasterized. (b) Given the 3D registered input image $(\textbf{I}_{y_s})$ and target expression $(\textbf{y}_t)$, target expression is transferred to input image $(\tilde{\textbf{I}}_{y_t})$. Using semantic resampling on the synthesized images, a training set of balanced AU intensity labels is formed. (c) Obtained balanced training set is used to train a convolutional neural network to estimate AU intensity values.}

\label{fig:pipeline}
\end{figure*}

Semi-supervised approaches deal with partially annotated data. They aim to leverage the unlabelled data with the assumption that unlabelled data follow continuity or form cluster with the labeled data \cite{Zhao18}.  Wu et al. \cite{wu2015multi} used Restricted Boltzmann Machine to model the AU distribution using the annotated labels, which is used to train the AU classifiers with partially labeled data. Zeng et al. \cite{zeng2015confidence} trained a quasi-semi-supervised (QSS) classifier with virtual labels provided by the confident positive and negative classifiers, which separate easily identified positive and negative samples from all else, respectively. 
Niu et al. \cite{niu2019multi} proposed a semi-supervised co-training approach named as multi-label coregularization for AU recognition, which aims to improve AU recognition with abundant unlabeled face images and domain knowledge of AUs. 

Recent works focus on self-supervised approaches where the goal is to learn the discriminative representation from the massive amount of videos without annotations. Li et al. \cite{Li19} proposed a self-supervised learning framework named Twin-Cycle Autoencoder that disentangles the AU-related movements from the pose-related ones to learn AU representations from unlabelled videos. While the aforementioned approaches tackle the problems in the annotations, none of them aims to balance the distribution of AU intensities. For label balancing, upsampling approaches where the infrequent labels are selected multiple times \cite{li2017eac}\cite{zhang2018identity} or multi-label minority oversampling majority undersampling approach \cite{chu2019learning} have been used. Since resampling is done within the dataset, such balancing methods do not contribute additional semantic information about the infrequent label.

\textbf{GAN-based facial expression transfer:} Recently GANs have received attention to transfer facial expressions from a source subject to a target subject. Existing work on GAN-based facial expression transfer approaches focus on generating facial images with discrete emotions \cite{choi2018stargan}\cite{ding2018exprgan}, or the specified facial action units \cite{Pumarola_ijcv2019}\cite{liu2020region}. Some of the GAN-based approaches specifically aim to guide their models with the facial geometry information. Song et al. \cite{song2018geometry} proposed a Geometry-Guided Generative Adversarial Network (G2-GAN) which employs fiducial points as a controllable condition to guide facial texture synthesis with specific expression. Qiao et al. \cite{qiao2018geometry} applied contrastive learning in GAN to embed geometry information onto a semantic manifold in the latent space for facial expression transfer. Geng et al. \cite{geng20193d} combined 3DMMs and deep generative techniques in a single framework for fine-grained face manipulation. Yet, in these studies transfer was limited to either 2D representation or to individual frames.

\textbf{Synthetic data augmentation:} Some of the recently proposed methods utilized synthetic data for facial expression analysis. Abbasnejad et al.~\cite{Abbasnejad2017} pre-trained their model using synthetic face images and then fine-tuned it on real images. Zhu et al.~\cite{Zhu2018} proposed a data augmentation method using GAN to classify basic emotions. Kollias et al.~\cite{Kollias2020} proposed an approach using 3DMMs to synthesize facial affect: in terms of six basic emotions or in terms of valence and arousal. Unlike the existing methods, our approach is designed to generate a large AU-balanced dataset.

\section{Proposed Method}

Fig.~\ref{fig:pipeline} shows the pipeline of our approach. First we perform dense 3D registration from 2D images. Then, we train a GANimation-based architecture with idiosyncratic loss to synthesize new facial expressions. Using semantic resampling, we obtain a balanced distribution of AU intensity labels. Finally, using the balanced synthetic expression database, we train a convolutional neural network architecture for AU intensity estimation.

\subsection{3D face registration}
\label{sec:3Dregistration}

We normalize videos using PRNet~\cite{feng2018prn}, a face alignment software that accomplishes dense 3D registration from 2D images without requiring person-specific training. PRNet uses an encoder-decoder architecture containing convolutional layers and residual blocks to jointly perform facial landmark alignment and 3D facial structure reconstruction. This architecture learns a mapping from an RGB image to UV position map (a 2D image representation of 3D coordinates in UV space keeping the position and semantic information). By learning the position map, it is possible to directly regress the complete 3D structure along with semantic meaning from a single image. Using PRNet, we obtain the dense 3D mesh of the face in a frontal view and texture information. Then, we map the texture to 3D mesh and rasterize it to 2D image of size 224 x 224.

\subsection{Facial Expression Generation Architecture}
\label{sec:Generation}
We build upon GANimation~\cite{Pumarola_ijcv2019} framework to synthesize novel facial expressions. First we map AU intensity labels (0 to E-level) to values in range [0,1]. Given a 3D registered source image $\textbf{I}_{{y_s}}$ with the AU intensity values $\textbf{y}_s = \{s_1, s_2, \ldots, s_n\}$, and target AU intensity values $\textbf{y}_t = \{t_1, t_2, \ldots, t_n\}$, we synthesize $\tilde{\textbf{I}}_{{y_t}}$. With our architecture, we aim to minimize these following terms:

\noindent \textbf{Image Adversarial Loss:} In order to obtain realistic synthesized images and ensure that the distribution of the generated images are similar to the distribution of the training images, we use image adversarial loss. Let $\mathcal{P}_s$ be the data distribution of the source image, $\mathcal{P}_{\tilde{I}}$ be the random interpolation distribution, and $\lambda_{gp}$ be the penalty loss. Then we can write the image adversarial loss $\mathcal{L}_{adv}(G,D_{adv}, \textbf{I}_{y_s}, \textbf{y}_t)$ as follows:

\begin{dmath}
    \mathbb{E}_{\textbf{I}_{y_s} \sim \mathcal{P}_s}[D_{adv}(G(\textbf{I}_{y_s}|\textbf{y}_t))]
    - \mathbb{E}_{\textbf{I}_{y_s} \sim \mathcal{P}_s}[D_{adv}(\textbf{I}_{y_s})] 
    + \lambda_{gp} \mathbb{E}_{\tilde{I} \sim \mathcal{P}_{\tilde{I}}}[(\| \nabla_{\tilde I} D_{adv}(\tilde I)\|_2 -1)^2]
\end{dmath}
where $G$ denotes generator and $D_{adv}$ denotes adversarial discriminator.

\noindent \textbf{Conditional Expression Loss:} In order to enforce $G$ to synthesize images containing the target expression $\textbf{y}_t$, we use the following loss $\mathcal{L}_{exp}(G,D_{exp}, \textbf{I}_{y_s}, \textbf{y}_s, \textbf{y}_t)$ to minimize the distance between AU intensities of the source and target images:

\begin{equation}
    \mathbb{E}_{\textbf{I}_{y_s} \sim \mathcal{P}_s}[\| D_{exp}(G(\textbf{I}_{y_s}|\textbf{y}_t)) - \textbf{y}_t\|_2^2] + 
    \mathbb{E}_{\textbf{I}_{y_s} \sim \mathcal{P}_s}[\| D_{exp}(\textbf{I}_{y_s}) - \textbf{y}_s\|_2^2]
\end{equation}
where $D_{exp}$ denotes discriminator for expression.

\noindent \textbf{Identity Loss:} We aim to guarantee that the face in both the input and output images belong to the same person. We use this cycle-consistency loss to penalize the difference between the original image $\textbf{I}_{y_s}$ and its reconstruction $\textbf{I}_{y_t}$.

\begin{equation}
    \mathcal{L}_{idt}(G, \textbf{I}_{y_s}, y_s, y_t) = \mathbb{E}_{\textbf{I}_{y_s} \sim \mathcal{P}_s}[\|G(G(\textbf{I}_{y_s}|{y_t})|{y_s}) - \textbf{I}_{y_s}\|_1]
\end{equation}

\noindent \textbf{Idiosyncratic Loss:} With the GANimation architecture, we can transfer the AU intensity values $\textbf{y}_t$ of a target image $\textbf{J}_{y_t}$ to the source image $\textbf{I}_{y_s}$ to synthesize  $\tilde{\textbf{I}}_{y_s}$. When the identity of source $(I)$ and target $(J)$ images are the same, then we can minimize the difference between $\textbf{I}_{y_t}$ and $\tilde{\textbf{I}}_{y_t}$ to ensure that both expression and identity of the synthesized image are the same as the target image. Idiosyncratic loss can be defined as:

\begin{equation}
    \mathcal{L}_{ids}(G, \textbf{I}_{y_s}, y_s, y_t) = \mathbb{E}_{\textbf{I}_{y_s} \sim \mathcal{P}_s}[\|(G(\textbf{I}_{y_s}|{y_t}) - \textbf{I}_{y_t}\|_1]
\end{equation}

\noindent \textbf{Final Loss:} We obtain our final loss by combining of the mentioned individual losses as follows:

\begin{equation}
    \mathcal{L} = \lambda_{adv} \mathcal{L}_{adv} + \lambda_{exp} \mathcal{L}_{exp} + \lambda_{idt} \mathcal{L}_{idt} + \lambda_{ids} \mathcal{L}_{ids}
\end{equation}
where $\lambda_{adv}$, $\lambda_{exp}$, $\lambda_{idt}$, and $\lambda_{ids}$ are the hyperparameters used to adjust the importance of different components.

\subsection{AU intensity estimation}

After we synthesize new expressions for the individuals in our database, we perform semantic resampling and create a training set having balanced AU intensity labels for each AU. Then we train convolutional neural networks (VGG16) using the balanced synthetic training set. During test time, we obtain AU intensity outputs of each estimator. 

\section{Experiments}
\label{sec:experiments}

\subsection{Datasets}

In all of our experiments, we used three facial expression datasets. For training the generator and evaluating within domain performance, we used the widely accepted 2017 Facial Expression Recognition Benchmark (FERA 2017) ~\cite{Valstar17}. For generating out-of-domain samples, we used the high resolution images from MultiPIE~\cite{gross2010multipie}. To evaluate the generalizability of our AU classifiers to another domain, we used the Denver Intensity of Spontaneous Facial Action Database (DISFA)~\cite{Mavadati13}. 
\newline

\noindent \textbf{FERA 2017:} 
The FERA 2017 Challenge was the first to provide a common protocol with which to compare approaches to detection of AU occurrence and AU intensity robust to pose variation. FERA 2017 provided synthesized face images with 9 head poses as shown in Fig.\ref{fig:non_frontal_view}. The training set is based on the BP4D database~\cite{Zhang14}, which includes digital videos of 41 participants. The development and test sets are derived from BP4D+~\cite{Zhang16} and include digital videos of 20 and 30 participants, respectively. FERA 2017 presented two sub-challenges: occurrence detection and intensity estimation. For the former 10 AUs were labelled; for the latter, 7 AUs were labelled. For our experiments, 7 AUs for intensity estimation were used.
\newline

\noindent \textbf{MultiPIE:}
The MultiPIE databset contains images of 337 people recorded in up to four sessions over the span of five months. Subjects were imaged under 15 view points and 19 illumination conditions while displaying a range of facial expressions. In addition, high resolution frontal images were acquired as well. For synthesis, we used the high resolution images only.
\newline

\noindent \textbf{DISFA:}
The DISFA dataset contains videos of 27 adult subjects (12 women, 15 men). It is manually annotated for AU intensity from 0 to E-level. Participants watched a video clip
consisting of 9 segments intended to elicit a range of facial
expressions of emotion.

\subsection{Experimental Setup}

In this section we describe the experimental setup for the generator network and the classifier.

\subsubsection{Generator Training}
In our experiments, we used the FERA 2017 dataset~\cite{Valstar17} to train facial expression generation models. The dataset consists of Train, Valid, and Test partition. We used Train partition only to train our model. All the images were resized to 224 $\times$ 224 pixels to match the receptive field of our AU estimation model (VGG16).

In all of our experiments, we used a GANimation ~\cite{Pumarola_ijcv2019} replicate implementation\footnote{https://github.com/donydchen/ganimation\_replicate}. We modified the loss function with the idiosyncratic constraint as described in the previous section.

\begin{figure*}[!htbp]
\begin{center}
\includegraphics[width=.60\linewidth]{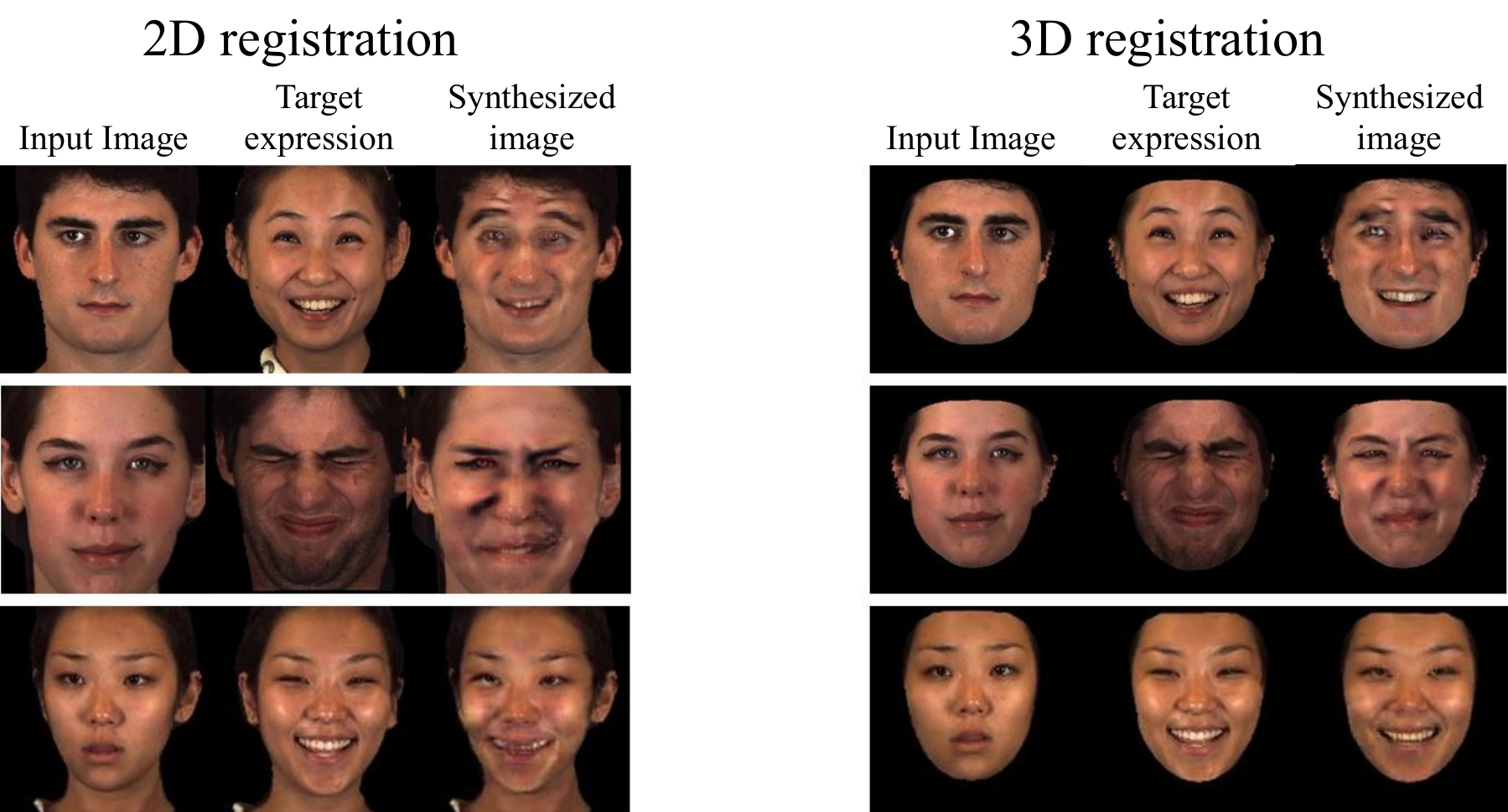}
\end{center}
\caption{Examples of 2D and 3D normalization. Note the higher image quality when 3D normalization was used.}
\label{fig:2dand3dregistratioin}
\end{figure*}

\subsubsection{Classifier Training}
\label{sec:dataset_au_estimation}

For the baseline experiments, we used the training partition of the FERA 2017 dataset~\cite{Valstar17} to train AU classifiers and used the test partition to test them. To create a balanced training set and compare methods, 5,000 frames were selected for each AU intensity. In the case of real images, we randomly selected 5,000 images from the six intensity classes (not present, and A to E levels). We down-sampled the majority classes and up-sampled the minority classes to reach this number. In the case of synthesized images, we first selected 5,000 pairs of input images and target AU labels for each intensity and each AU. Then we synthesized 5,000 images for these pairs.
We selected 5,000 images according to experimental results by Niinuma et al.~\cite{Niinuma19}. They analyzed the influence of training set size on FERA17, and showed that the training set sized have minor influence on the performance: score peaked at 5,000 images, after that performance plateaued.
During the selection input images were selected only from frames that did not have the target AU of interest. For example, when we synthesized images for E level intensity of AU1, we randomly selected 5,000 input images that did not have AU1 present and 5,000 target AU labels having E level intensity for AU1. We employ this strategy for two main reasons. First, while the generator can add realistic facial deformations (like wrinkles and bulges) to neutral faces, it oftentimes fails in removing those. Therefore, to acquire higher quality synthesized images, starting from neutral frame is preferable. Second, since the AU labels are sparse, there are many more frames in the dataset where the target AU is not present. This way we can obtain a variety of synthesized images for each facial expression.

We selected a VGG16 network pre-trained on ImageNet for the baseline architecture for AU estimation. Previous studies found this combination preferable for AU coding \cite{Niinuma19}. We replaced the final layer of the network with a 6-length one-hot representation, and fine-tuned VGG16 network from the third convolutional layer. Dropout rate was set to $0.5$, and Adam optimizer was used with $LR=5 \times 10^{-5}$ as suggested in ~\cite{Niinuma19}.

\subsection{Synthetic vs Real Expressions under 2D vs 3D Alignment}

In this set of experiments we studied how two main components affect the performance of the whole system. 

First, we were interested in the effect of face alignment on the synthesis and AU recognition performance. We explored both 2D and 3D alignment. 2D normalization treats the face as 2D object. That assumption is reasonable, as long as there is no head movement present. As soon as head orientation deviates from frontal, one expects the classifier’s ability to measure expressions to degrade. On the other hand, 3D normalization should be able to preserve semantic correspondences of the different facial regions across poses, and result in higher performance. For 2D alignment, we applied Procrustes analysis between 68 landmarks provided by the dlib face tracker\cite{dlib09} on the frames and a frontal template. For 3D normalization we used the method described in Sec.~\ref{sec:3Dregistration}.
Note that there are not significant differences between shallow and deep approaches in terms of facial alignment~\cite{Jeni2016}\cite{Sagonas2016}. In the 300-W Challenge~\cite{Sagonas2016}, the two top methods are a cascade regressor and a CNN approaches. The dlib face tracker is an implementation of a decision tree based cascade regressor~\cite{Kazemi2014}.

Second, we were interested how synthetic expressions would affect the classification performance. We compared multi-label minority oversampling $\&$ majority undersampling with the proposed, completely synthetic expression generation. Both methods balance the skewed distributions of AUs, but while the first one can not produce more varied minority samples, the latter one can. As mentioned in Sec.~\ref{sec:dataset_au_estimation}, we used 5,000 images for each intensity each AU for both real and synthesized image settings.

\begin{table}[!htbp]
\caption{Comparison of synthetic vs real expressions under 2D vs 3D alignment. Scores are Inter-rater reliability (ICC) of AU intensity level estimation. The Images row shows which is used to train classifiers: Real or Synthetic. All of the classifiers were tested on the real test dataset. The same Registration (2D or 3D) were applied to both train and test datasets.}
\label{tb:comp_baselines}
\centering
\begin{tabular}{|c|cccc|}
\hline

\hline
Registration & 2D & 2D & 3D & 3D\\
Images & Real & Synthetic & Real & Synthetic \\
\hline
AU1&\bf{0.431} &0.336 &0.343 &0.381\\
AU4&0.223 &0.116 &\bf{0.260} &0.219\\
AU6&0.796 &0.790 &0.751 &\bf{0.804}\\
AU10&0.777 &\bf{0.812} &0.785 &0.773\\
AU12&0.801 &0.792 &\bf{0.806} &0.795\\
AU14&0.118 &0.238 &0.084 &\bf{0.244}\\
AU17&0.395 &0.374 &0.391 &\bf{0.461}\\
\hline
Mean&0.506 &0.494 &0.489 &\bf{0.525}\\
\hline
\end{tabular}
\vspace{-1em}
\end{table}

Fig. \ref{fig:2dand3dregistratioin} shows examples of synthetic expressions generating using 2D and 3D alignment. 3D alignment results in less ghosting and other texture artifacts and provide higher quality images. Inter-rater reliability (ICC) results of AU classification under these four different conditions are shown in Table.~\ref{tb:comp_baselines}. 
3D registration with synthetic re-sampling outperformed the other three conditions.

\begin{figure*}[!h]
\begin{center}
\includegraphics[width=.80\linewidth]{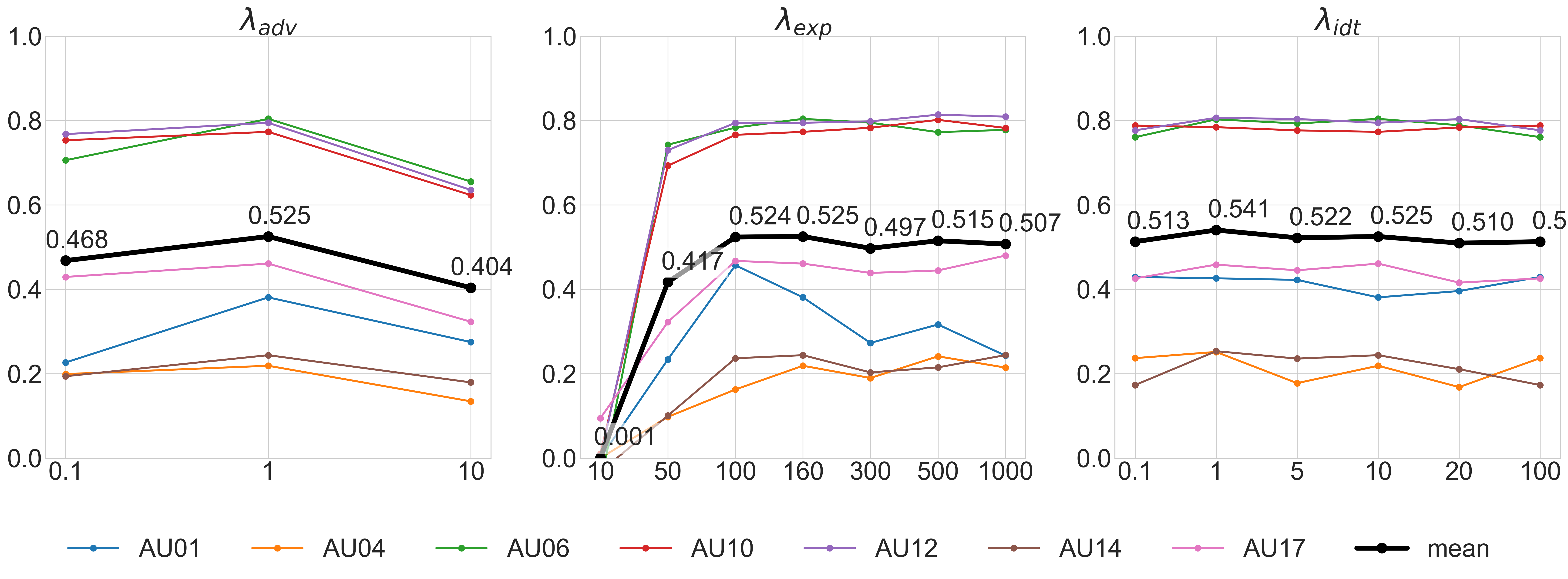}
\end{center}
\caption{Influence of parameter values on ICC for intensity estimation }
\label{fig:params}
\end{figure*}

In these experiments we conducted a parameter search to find the optimal values of the generator for classification. We varied parameters that control the contribution of the adversarial loss ($\lambda_{adv}$) , conditional expression loss ($\lambda_{exp}$), and the identity loss ($\lambda_{idt}$). 
In our baseline configuration, we used 1.0 for $\lambda_{adv}$, 160 for $\lambda_{exp}$, and 10 for $\lambda_{idt}$.
Fig.~\ref{fig:params} shows the impact of parameter values on the intensity estimation for different AUs. $\lambda_{idt}$ does not affect the classification significantly. For both $\lambda_{exp}$ and $\lambda_{adv}$, we selected the global optimal values. Pumarola et al. introduced an attention mask and a color transformation term in the lost function to prevent the attention mask saturation~\cite{Pumarola_ijcv2019}. In our experiments we did not observe this saturation effect and removed these terms from the loss function.

\begin{figure*}[!h]
\begin{center}
\includegraphics[width=.78\linewidth]{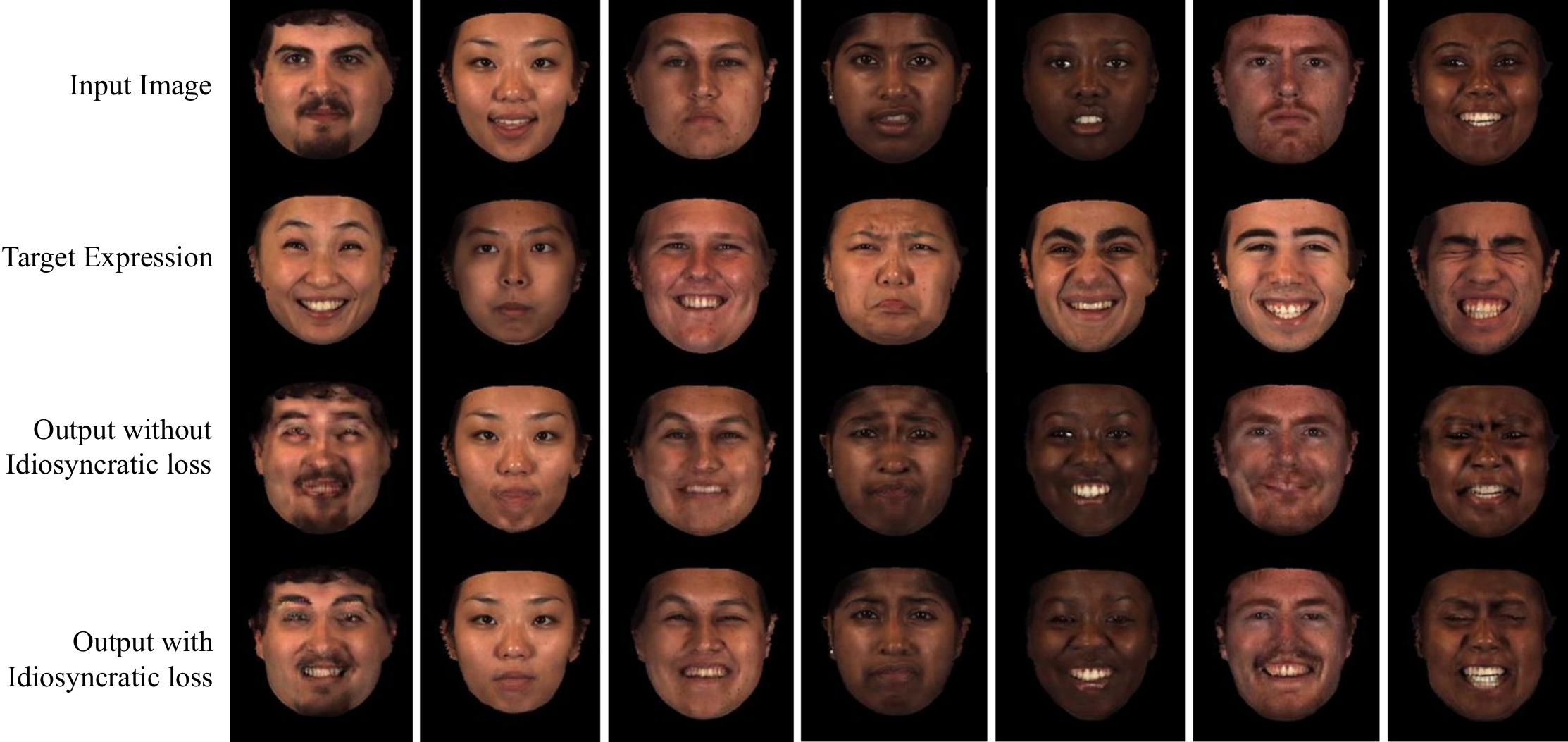}
\end{center}
\caption{Comparison without and with idiosyncratic loss}
\label{fig:target_loss}
\end{figure*}

\subsection{The Effect of Idiosyncratic Loss}

3D alignment improves the performance, image quality is still low in some cases especially when target expression has a high AU intensity. To mitigate the problem, we introduced a new idiosyncratic term in the loss function. FERA 2017 datasets include many images with different facial expressions for each subject. Idiosyncratic loss utilizes this feature of the datasets. Fig.~\ref{fig:target_loss} shows some of the examples ($\lambda_{ids} = 1$). The ICC result with the new function is 0.523. 
To calculate the ICC result, 3D synthetic train images were used to train classifiers, and 3D real test images were used to test the classifiers.
Compared with the ICC without it (0.525), it does not improve the ICC, but we confirmed that it improves the image quality.
The Frechet Inception Distance (FID)~\cite{Heusel2017FID} for the synthetic dataset with idiosyncratic loss (4.88) is better than the one without idiosyncratic loss (5.94). Note that a lower FID is better.

\subsection{Temporal Normalization and Comparison with State of the Art}

Results from the previous experiment suggest that precise spatial alignment improves the performance. AUs are temporal, we decided to test the best method under different temporal normalization. 
We compared two methods that enhances the temporal aspect of the AUs: AU0 normalization and mean texture normalization. AU0 normalization computes the appearance differences between the actual frame and a neutral frame. A neutral frame is not necessary available in real life conditions, but we can assume that we have multiple frames from a single person. In personal mean texture normalization we calculate the average appearance of a person and then calculate the differences between each frame and the mean texture. This step minimizes individual differences in the appearance space.

\begin{table*}[!htbp]
\caption{Comparison with state-of-the-art methods. Reported scores are Inter-rater reliability (ICC) on frontal views only. The results for No norm is the same with the ones for 3D Synthetic in Table~\ref{tb:comp_baselines}.}
\label{tb:comp_existingwork}
\centering
\begin{tabular}{|c|ccccc|ccc|}
\hline
& Valstar & Amirian & Batista  &   Zhou & Niinuma & & Ours &  \\ 
&  et al.~\cite{Valstar17} &  et al.~\cite{Amirian17} & et al.~\cite{Batista17} &  et al.~\cite{Zhou17}  &   et al.~\cite{Niinuma19} & No norm & AU0 norm & Mean norm
\\

\hline
AU1&0.025&0.270 &0.311 &0.286 &0.433 &0.381 &0.539 &\bf{0.613}\\
AU4&0.003&0.074 &0.098 &0.130 &0.281 &0.219 &0.361 &\bf{0.409}\\
AU6&0.616&0.644 &0.721 &0.625 &0.786 &\bf{0.804} &0.764 &0.779\\
AU10&0.662&0.733 &0.741 &0.739 &0.768 &0.773 &\bf{0.787} &0.757\\
AU12&0.709&0.745 &0.754 &\bf{0.822} &0.812 &0.795 &0.792 &0.794\\
AU14&0.066&0.030 &0.127 &0.075& 0.153 &\bf{0.244} &0.114 &0.170\\
AU17&0.015&0.271 &0.252 &0.342 &0.382 &0.461 &0.288  &\bf{0.465}\\
\hline
Mean&0.299&0.395 &0.429 &0.431 &0.516 &0.525 &0.521 &\bf{0.570}\\
\hline
\end{tabular}
\vspace{-1em}
\end{table*}

The last three columns of Table.~\ref{tb:comp_existingwork} show the results. Personal mean texture normalization shows the best results. On average, there is no gain with AU0 normalization, however, individual AU level differences are significant. While AU0 normalization shows better results for AU1 and AU4, it shows  worse results for AU14 and AU17.

Table.~\ref{tb:comp_existingwork} shows the comparison with other state-of-the-art methods using only the frontal poses. We report Inter-rater reliability (ICC) that is the standard metric of the FERA 2017 benchmark. Our method (with or without temporal enhancement) outperforms all other methods. For a fair comparison, we compared our approach with existing methods' results on frontal view of the test partition.

\begin{figure*}[!htbp]
\begin{center}
\includegraphics[width=.64\linewidth]{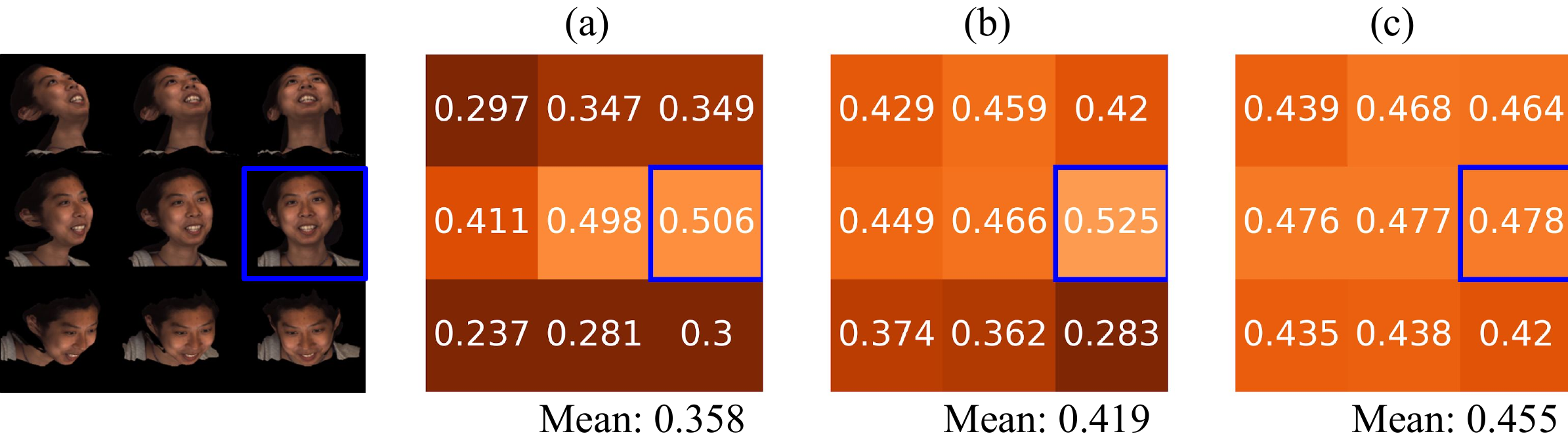}
\end{center}
\caption{ICC for Test partition with non-frontal view.}
\label{fig:non_frontal_view}
\end{figure*}

\begin{table*}[]
\caption{Cross domain ICC performance. (Synthetic training set $\rightarrow$ Real test set)}
\label{table:corss-domain}
\centering
\begin{tabular}{|c|ccc|}
\hline
     & MultiPIE$\rightarrow$FERA & FERA$\rightarrow$DISFA & MultiPIE$\rightarrow$DISFA \\
\hline
AU01 & 0.311                       & 0.314                    & 0.418                        \\
AU04 & 0.202                       & 0.400                    & 0.541                        \\
AU06 & 0.786                       & 0.573                    & 0.524                        \\
AU10 & 0.726                       & -                        & -                            \\
AU12 & 0.792                       & 0.748                    & 0.698                        \\
AU14 & 0.168                       & -                        & -                            \\
AU17 & 0.365                       & 0.373                    & 0.290                        \\
\hline
Mean & 0.479                       & 0.482                    & 0.494   \\
\hline     

\end{tabular}
\vspace{-1em}
\end{table*}

\subsection{Experiment with Non-frontal Poses}

Encouraged by the results of the previous experiment, we decided to evaluate the method's ability to generalize to unseen poses. FERA 2017 has nine different poses, we report the performance on all of these using the Test partition. We investigated three scenarios:

\begin{enumerate}
\item[(a)] \textbf{real 2D normalized $\rightarrow$ real 2D normalized.} For a baseline, we used real images with 2D alignment for training, and evaluated performance on 2D normalized real images from the testing set.
\item[(b)] \textbf{synthetic 3D normalized $\rightarrow$ real 3D normalized} We trained on synthetic 3D normalized images, and tested on 3D normalized real images from the test set. We applied the 3D normalization procedure described in Sec.~\ref{sec:3Dregistration} to each test image with non-frontal pose. Self-occluded facial parts were filled with black color during the rasterization step.
\item[(c)] \textbf{3D augmented synthetic $\rightarrow$ real 2D normalized} We synthesized 3D meshes and rotated them into the nine standard orientations found in FERA 2017. We randomly selected 500 images for each intensity, AUs, poses. The total number of images for each intensity each AU is 4,500 while 5,000 images are selected for scenarios (a) and (b). We tested the system on 2D aligned test images.
\end{enumerate}

Fig.~\ref{fig:non_frontal_view} shows the results. While (a) and (b) show low ICCs when face poses are largely different from frontal views, the performance drop for (c) is much smaller. The results show that our synthesized images with 3D registration are also effective to non-frontal views by recreating non-frontal view images from the synthesized images. Note that: 1) the approaches in Table.~\ref{fig:non_frontal_view} use frontal view images only to train models while the methods in FERA17 challenges used images with all 9 poses to train models., 2) the reason why the performance on frontal view for (c) is worse than (a) and (b) is that only 500 frontal view images for each intensity each AU are used to train.

\subsection{Cross-domain Experiments}
We have learned from the previous experiments that our AU classifiers can perform well when trained and tested within the same domain. To evaluate the generalizability of our approach to unseen domains (both in generating expressions and evaluating classifiers), we conducted two sets of experiments. 

First, we were interested in how generating out-of-domain samples would affect the performance on the FERA 2017 Test partition. In this case we trained the generator network on FERA 2017 Train partition, but we synthesized new expressions using high resolution frontal images present in the MultiPIE dataset. We selected 921 MultiPIE images and generated 5 target expressions each, resulting in 4,605 images for each intensity and each AU. We trained the classifier on these images and tested the performance on FERA 2017 Test partition.

In the second cross-domain experiment we evaluated out-of-domain classification. Here classifiers were trained either on synthetic expressions generated from FERA 2017 or synthetic expressions generated from MultiPIE, and they were tested on DISFA. The DISFA dataset differs in imaging condition and type of AU coding: Context is not social in DISFA while in FERA subjects are interacting with the experimenter, and in DISFA, the base rates of most AUs is very low and limited to what occurs in a film-watching paradigm~\cite{Ertugrul2020}.

In all of these experiments, FERA17 Train partition was used to train facial expression generation models, and 3D normalization was applied to each image. The whole dataset of DISFA was used to test the models.

Table.~\ref{table:corss-domain} shows the results. Performance of models trained on synthesized MultiPIE expressions (0.479) is lower than the one trained on synthesized FERA 2017 expression (0.525), but there is only 1\% difference with the one trained on real FERA17 expressions (0.489). The results on DISFA shows that the result trained on synthesized MultiPIE expressions (0.494) is slightly better than the one trained on synthesized FERA17 expressions (0.482).

\vspace{-0.5em}
\section{Conclusion}

We have proposed a generative approach that achieves 3D geometry based AU manipulation with idiosyncratic loss to synthesize facial expressions. With the semantic resampling, our approach provides a balanced distribution of AU intensity labels, which is crucial to train AU intensity estimators. We have shown that using the balanced synthetic set for training performs better than using the real training dataset on the same test set. Generating expressions using the 3D registered facial images gives better AU intensity estimation performance compared to using 2D registered ones. Moreover, our proposed idiosyncratic loss has improved the visual quality of the outputs. Cross-pose and cross-domain results reveal that classifiers trained on our synthesized images are also effective to non-frontal views and to unseen domains. 

\section{Acknowledgments}

This research was supported in part by Fujitsu Laboratories, NIH awards NS100549 and MH096951, and NSF
award CNS-1629716.

{\small
\bibliographystyle{ieee_fullname}
\bibliography{egbib}
}

\include{egpaper_supp}

\end{document}

%% file: egpaper_supp.tex
\onecolumn

\section*{S1 Supplemental material}
\subsection*{S1.1 Experiments for models trained on combined datasets}

In the paper, we report results for models trained on single datasets. In this supplementary material, we report results for models trained on combined datasets. Table~\ref{tb:FERA_results} and ~\ref{tb:DISFA_results} show the results for FERA Test partition and DISFA, respectively. The first, second and forth columns show the results for models trained on single datasets, and the third, fifth and sixth columns show results for models trained on combined datasets. As shown in the tables, combined datasets show slightly better results.

\begin{table*}[!htbp]
 \renewcommand\thetable{S1}
\caption{ICC comparison of models trained on single vs combined datasets for FERA 2017 Test partition. Training size is the number of images per intensity per AU used to train models.}
 \label{tb:FERA_results}
 \centering
\begin{tabular}{|c|ccc:ccc|}
 \hline
& FERA real & \begin{tabular}[c]{@{}l@{}}FERA\\     synthetic\end{tabular} & \begin{tabular}[c]{@{}l@{}}FERA real\\     +\\     FERA\\     synthetic\end{tabular} & \begin{tabular}[c]{@{}l@{}}MultiPIE\\     synthetic\end{tabular} & \begin{tabular}[c]{@{}l@{}}FERA real\\     +\\     MultiPIE\\     synthetic\end{tabular} & \begin{tabular}[c]{@{}l@{}}FERA\\     synthetic\\     +\\     MultiPIE\\     synthetic\end{tabular}\\
\hline
Training size& 5,000 &5,000 &10,000 & 4,605 & 9,605 & 9,605 \\ 
\hline
AU01  & 0.343      & 0.381           & 0.446 & 0.311                                                                                   & 0.315 & \bf{0.457}                                                                                                                                                         \\
AU04  & 0.260      & 0.219          & 0.324& 0.202                                                                                    & \bf{0.342} & 0.288                                                                                                                                                         \\
AU06  & 0.751      & \bf{0.804}              & 0.794          & 0.786                                                                       & 0.784 & 0.801                                                                                                                                                        \\
AU10  & \bf{0.785}      & 0.773              & 0.772          & 0.726                                                                      & 0.760 & 0.763                                                                                                                                                         \\
AU12  & 0.806      & 0.795              & 0.800          & 0.792                                                                      & \bf{0.807} & 0.801                                                                                                                                                         \\
AU14  & 0.084      & \bf{0.244}              & 0.171          & 0.168                                                                      & 0.116    & 0.235                                                                                                                                                     \\
AU17  & 0.391      & \bf{0.461}              & 0.433          & 0.365                                                                      & 0.408 & 0.436                                                                                                                                                         \\
\hline
Mean  & 0.489      & 0.525              & 0.534          & 0.479                                                                       & 0.505 & \bf{0.540}                                                                                                                                                         \\  
\hline
\end{tabular}
\end{table*}

\begin{table*}[!htbp]
 \renewcommand\thetable{S2}
\caption{ICC comparison of models trained on single vs combined datasets for DISFA. Training size is the number of images per intensity per AU used to train models.}
 \label{tb:DISFA_results}
 \centering
\begin{tabular}{|c|ccc:ccc|}
 \hline
 & FERA real & \begin{tabular}[c]{@{}l@{}}FERA\\     synthetic\end{tabular} & \begin{tabular}[c]{@{}l@{}}FERA real\\     +\\     FERA\\     synthetic\end{tabular} &
 \begin{tabular}[c]{@{}l@{}}MultiPIE\\     synthetic\end{tabular} & 
 \begin{tabular}[c]{@{}l@{}}FERA real\\     +\\     MultiPIE\\     synthetic\end{tabular} & \begin{tabular}[c]{@{}l@{}}FERA\\     synthetic\\     +\\     MultiPIE\\     synthetic\end{tabular}  \\
\hline
Training size& 5,000 &5,000 &10,000 & 4,605 & 9,605 & 9,605 \\ 
\hline
AU01  & 0.394     & 0.314                                                            & 0.365                                                        & 0.418                                                                                & \bf{0.470} & 0.346                                                                                                                                                                                   \\
AU04  & \bf{0.634}     & 0.400                                                            & 0.571                                                        & 0.541                                                                                & 0.544 & 0.402                                                                                                                                                                                   \\
AU06  & 0.404     & 0.573                                                            & 0.507                                                        & 0.524                                                                                & 0.496 & \bf{0.576}                                                                                                                                                                                   \\
AU12  & 0.750     & 0.748                                                            & 0.723                                                        & 0.698                                                                                & 0.749 & \bf{0.762}                                                                                                                                                                                   \\
AU17  & 0.293     & 0.373                                                            & 0.296                                                        & 0.290                                                                                & 0.377 & \bf{0.390}                                                                                                                                                                                   \\
\hline
Mean  & 0.495     & 0.482                                                            & 0.493                                                        & 0.494                                                                                & \bf{0.527} & 0.495                                                                                                                                                                                   \\
\hline                                                                                
\end{tabular}
\end{table*}

\clearpage

\subsection*{S1.2 Comparison without and with idiosyncratic loss}

Table~\ref{tb:comp_idiosyncratic} shows the results without and with idiosyncratic loss ($\lambda_{ids} = 1.0)$. It does not improve the ICC, but we confirmed that it improves the image quality as mentioned in Sec. 4.4.

\subsection*{S1.3 Synthetic MultiPIE images}
\begin{figure*}[!htbp]
 \renewcommand\thefigure{S1}
\begin{center}
\includegraphics[width=.78\linewidth]{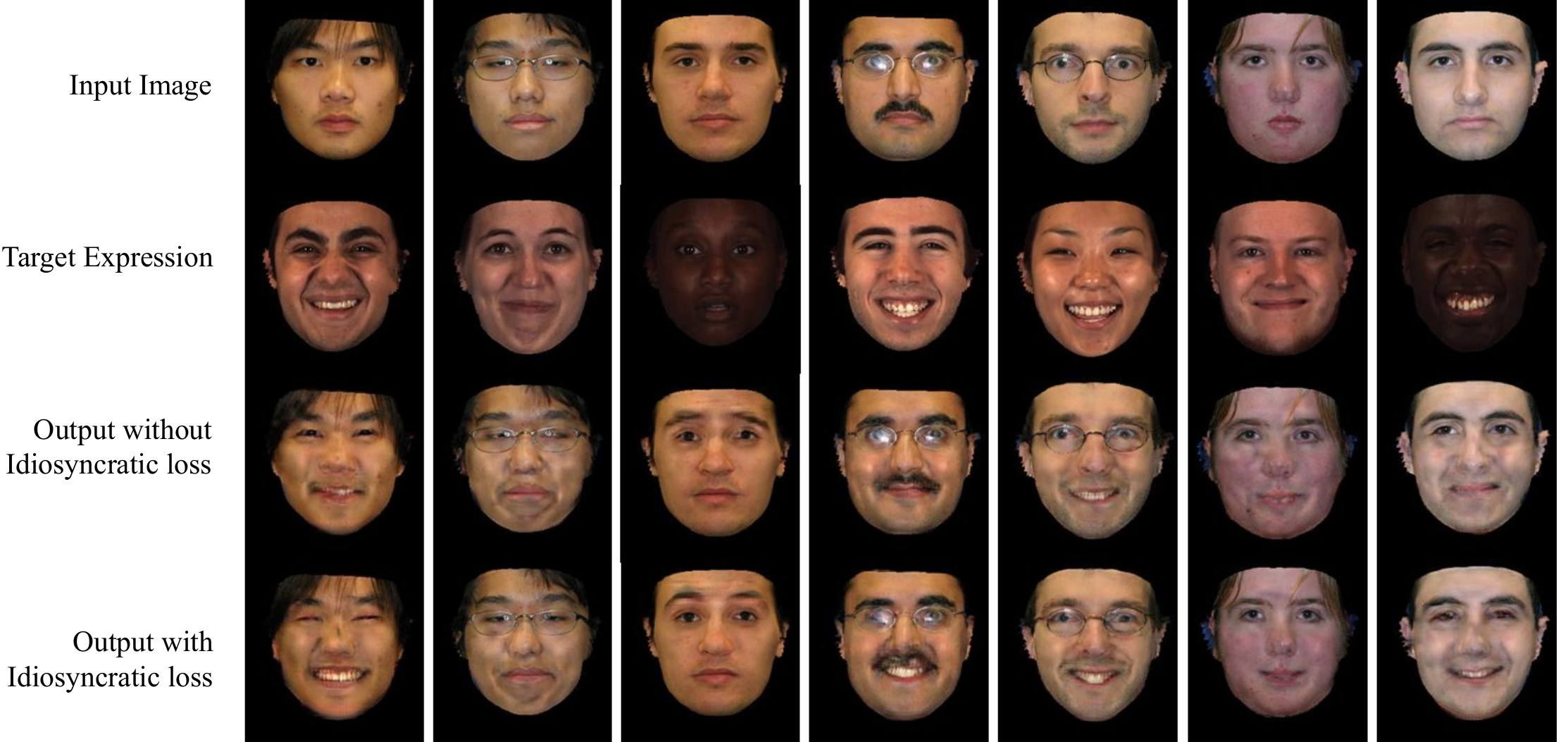}
\end{center}
\caption{Comparison without and with idiosyncratic loss on MultiPIE.}
\label{fig:multipie}
\end{figure*}

Fig.~\ref{fig:multipie} shows some examples of MultiPIE synthesized expressions without and with idiosyncratic loss.

\begin{table*}[!htbp]
 \renewcommand\thetable{S3}
\caption{ICC for intensity estimation without and with idiosyncratic loss on FERA 2017 Test partition.}
\label{tb:comp_idiosyncratic}
\centering
\begin{tabular}{|c|cc|}
\hline
& Without idiosyncratic loss &With idiosyncratic loss\\
\hline
AU1 &\bf{0.381} &0.350\\
AU4 &0.219 &\bf{0.302}\\
AU6 &\bf{0.804} &0.802\\
AU10 &0.773 &\bf{0.780}\\
AU12 &\bf{0.795} &0.793\\
AU14 &\bf{0.244} &0.181\\
AU17 &\bf{0.461} &0.452\\
\hline
Mean &\bf{0.525} &0.523 \\
\hline
\end{tabular}
\end{table*}

\clearpage

\subsection*{S1.4 Comparison of GAN architectures}

To examine the influence of GAN architectures, we coducted experiments for GANimation, StarGAN and GANimation internal classifier $D_{exp}$. Table~\ref{tb:comp_GAN} indicates that GANimation and StarGAN show almost the same performance, but GANimation internal classifier shows worse than the others.

\subsection*{S1.5 Cross-domain experiments on UNBC Pain}
Table~\ref{tb:cross_domain_pain} shows the cross domain results for UNBC Pain. Classifiers were trained either on synthetic expressions generated from FERA 2017 or synthetic expressions generated from MultiPIE, and they were tested on UNBC Pain.

\begin{table*}[!htbp]
 \renewcommand\thetable{S4}
\caption{ICC comparison for GAN architectures.}
\label{tb:comp_GAN}
\centering
\begin{tabular}{|c|ccc|}
\hline
& \multirow{2}{*}{GANimation} &\multirow{2}{*}{StarGAN} &GANimation\\
&& & internal classifier\\
\hline
AU1 &\bf{0.381} &0.367 &0.380\\
AU4 &0.219 &\bf{0.259} &0.065\\
AU6 &\bf{0.804} &0.793 &0.712\\
AU10 &0.773 &\bf{0.788} &0.743\\
AU12 &0.795 &\bf{0.807} &0.793\\
AU14 &\bf{0.244} &0.199 &0.123\\
AU17 &\bf{0.461} &0.451 &0.364\\
\hline
Mean &\bf{0.525} &0.523 & 0.454\\
\hline
\end{tabular}
\end{table*}

\begin{table*}[!htbp]
 \renewcommand\thetable{S5}
\centering
\caption{Cross domain ICC performance for UNBC Pain. (Synthetic training set $\rightarrow$ Real test set)}
\label{tb:cross_domain_pain}
\begin{tabular}{|c|cc|}
\hline
     & FERA$\rightarrow$UNBC Pain & MultiPIE$\rightarrow$UNBC Pain \\ \hline
AU04 & 0.130                      & 0.149                          \\
AU06 & 0.496                      & 0.434                          \\
AU10 & 0.034                      & 0.038                          \\
AU12 & 0.402                      & 0.367                          \\ \hline
Mean & 0.266                      & 0.247                          \\ \hline
\end{tabular}
\end{table*}